\documentclass[11pt,letterpaper]{article}
\usepackage{naaclhlt2016}
\usepackage{times}
\usepackage{latexsym}

\usepackage{url}
\usepackage{graphicx}
\usepackage{color}
\usepackage{multirow}
\usepackage{amsmath}
\usepackage{subfigure}
\usepackage{color}

\naaclfinalcopy % Uncomment this line for the final submission
\def\naaclpaperid{***} %  Enter the naacl Paper ID here

\title{Part-of-Speech Relevance Weights for Learning Word Embeddings}
\author{Quan Liu$^\dagger$, Zhen-Hua Ling$^\dagger$, Hui Jiang$^\ddagger$, Yu Hu$^\dagger$$^\S$\\
$^\dagger$ National Engineering Laboratory for Speech and Language Information Processing \\
University of Science and Technology of China, Hefei, Anhui, China \\
$^\ddagger$ Department of Electrical Engineering and Computer Science \\
York University,  4700 Keele Street, Toronto, Ontario, M3J 1P3, Canada \\
$^\S$ iFLYTEK Research, Hefei, China\\
\em emails: quanliu@mail.ustc.edu.cn, zhling@ustc.edu.cn \\
\em hj@cse.yorku.ca, yuhu@iflytek.com
}

\date{}

\begin{document}

\maketitle

\begin{abstract}
This paper proposes a model to learn word embeddings with weighted contexts based on part-of-speech (POS) relevance weights.
POS is a fundamental element in natural language.
However, state-of-the-art word embedding models fail to consider it.
This paper proposes to use position-dependent POS relevance weighting matrices to model the inherent syntactic relationship among words within a context window.
We utilize the POS relevance weights to model each word-context pairs during the word embedding training process.
The model proposed in this paper paper jointly optimizes word vectors and the POS relevance matrices.
Experiments conducted on popular word analogy and word similarity tasks all demonstrated the effectiveness of the proposed method.
\end{abstract}

\section{Introduction}

Word embedding that represents words into continuous vector space based on distributional hypothesis is an important research topic in the natural language processing community \cite{hinton1986distributed,turian2010word,mikolov2013efficient}.
State-of-the-art word embedding models include the neural language models \cite{bengio2003neural,mikolov2010recurrent}, the C\&W model \cite{collobert2011natural}, the continuous bag-of-word (CBOW) and Skip-gram word2vec models \cite{mikolov2013efficient}, and the GloVe model \cite{pennington2014glove}.
Word embedding techniques have been widely applied to various natural language processing tasks, including machine translation \cite{devlin2014fast,wuemnlp14}, sequence labelling \cite{collobert2011natural} and antonym selection \cite{xiaodan15}.
%% Distributional hypothesis
Under the framework of current word embedding models, word vectors are estimated based on the \textit{distributional hypothesis} \cite{harris1954distributional}.
%The distributional hypothesis assumes that words with a similar context tend to have a similar meanings.
%Therefore, most of the state-of-the-art models tend to define a window of words as context for model training.
 It has been found that the learned vectors could explicitly encode many linguistic regularities and patterns \cite{mikolov2013distributed}.
However, it is still inadequate to learn high-quality representations just relying on word level distributional information collected from text corpora \cite{faruqui15retrofit,quanliu15}.

%% Powered word embeddings
To improve the representation of word embeddings, some recent works have been proposed to incorporate various kinds of additional resources into the word representation learning framework.
Typically, some knowledge enhanced word embedding models tried to exploit lexical knowledge resources as semantic constraints for learning word embeddings \cite{yuimprove14,xu2014rc,faruqui15retrofit,quanliu15}.
In \newcite{levy2014depend}, they investigated to use syntactic contexts that were derived from automatically produced dependency parse-trees for word representation training.
Meanwhile, some people attempted to utilize multilingual parallel corpora to guide the word vector training process \cite{zhangjia2014,hermann2014,lu2015naacl}.
Nevertheless, all those works failed to consider the basic POS information in the training corpus.
%Since part-of-speech is a fundamental element in natural language processing system, %it is necessary to exploit its impacts on meaning representation.
%it may be beneficial to incorporate it into word representation

To exploit the effectiveness of POS information for word representations, this paper proposes a model to incorporate it into the training process of word embeddings.
POS tags capture syntactic roles of words and ignore much of their lexical information \cite{jelinek1990self}.
%Therefore, in order to utilize part-of-speech information for word embeddings, it is not a good choice to treat each POS tag as a specific semantic category like the work of sentiment specific word embedding.
This paper considers that, for a word-context pair, the collocation of their POS tags encodes their inherent syntactic relationships. This paper proposes to model the syntactic relationships of word-context pairs with a set of position-dependent POS relevance weighting matrices.
After that, all word-context pairs used for learning word embeddings are weighted by the POS relevance weights.
Finally, this paper proposes to learn the relevance weights and word embeddings jointly using the stochastic gradient descend (SGD) algorithm.
Experiments conducted on the popular word analogy and word similarity tasks have demonstrated that, by introducing POS relevance weight as a discriminative factor for context weighting, the quality of the learned word
vectors is significantly improved comparing to the baseline model.

\section{Related Work}
POS has been used in various natural language processing tasks such like language modeling \cite{heeman1998pos}, dependency parsing \cite{koo2008simple} and name entity recognition \cite{turian2010word}.
However, very few of them investigated to use POS information for distributed word representation.
The most related one was the method proposed by \newcite{qiu2014learning} which trained one representation vector for each
POS tag of a word and did not try to improve the representation ability of word embeddings.
In order to exploit POS information, this paper borrows the idea of language modeling from \cite{jelinek1990self,brown1992class,heeman1998pos} and proposes to use POS relevance weighting matrices to model each word-context pair for word embedding learning.
%To the best of our knowledge, this is the first work that incorporates part-of-speech relevance weights as context weighting factors for learning word embeddings.
%This is mainly because that part-of-speech is more abstract than words.
%Therefore, one recent work proposed by \newcite{qiu2014learning} tends to train one representation vector for each POS tag of a word.
%In order to exploit part-of-speech information, this paper borrows the idea of class-based language model \cite{brown1992class} and propose part-of-speech transition matrices to scale each word-context pair for word vector training.
%To the best of our knowledge, this is the first work that incorporates part-of-speech as context scaling factor for learning distributed word representations.

\section{The proposed model}
\label{sec:main}

In this section, we present the proposed model that employs POS information for learning word embeddings, called \textbf{PWE} hereafter.
Given a large training corpus, the POS information can be obtained efficiently by a state-of-the-art POS tagger.
Consider a sequence of words $S=\{w_{1},...,w_{N}\}$ with $N$ word tokens.
After conducting POS tagging, each word token $w_{i}$ is labelled as a specific POS tag $z_{i}$.
The corresponding word-POS pairs are denoted as $\langle w_{i}, z_{i}\rangle$.

\subsection{Main framework}
\label{ssec:scaling}
This paper proposes to incorporate POS information for learning word embeddings based
on the CBOW model \cite{mikolov2013efficient}.
We firstly review the objective function of a typical CBOW model, which is to maximize the log likelihood of each token given its contexts:
\begin{equation}\label{eq:cbow-obj}
  {\cal Q}_{\mathrm{cbow}}
= \frac{1}{T} \sum_{t=1}^{T}{ \log p(w_{t} | w_{t+c}^{t-c}) }
\end{equation}
where $c$ specifics the context window size, $T$ is the token number of the training corpus.
Meanwhile, the word prediction probability is
\begin{equation}\label{eq:prob}
  p(w_{t}|w_{t+c}^{t-c}) = \frac{\exp\left(\mathbf{w}_{t}^{(2)} \cdot \mathbf{v}_{t-c}^{t+c}\right)} {\sum_{k=1}^{V} {\exp\left(\mathbf{w}_{k}^{(2)} \cdot \mathbf{v}_{t-c}^{t+c}\right)} }
\end{equation}
where $\mathbf{v}_{t-c}^{t+c}$ is the context representation, $\mathbf{w}_{t}^{(2)}$ is the output vector of word $w_{t}$, $V$ is the vocabulary size.
In the CBOW model, the context representation is calculated by bag-of-word averaging
\begin{equation}\label{eq:bow-vec}
  \mathbf{v}_{t-c}^{t+c} = \sum_{-c \leq i \leq c, i \ne 0} \mathbf{w}_{t+i}^{(1)}
\end{equation}
where $\mathbf{w}_{t+i}^{(1)}$ is the word embedding for word $w_{t+i}$.

The CBOW model treats word-context pairs equally and fails to utilize the discrimination information encoded in their inherent syntactic relationships.
Motivated by this deficiency, this paper proposes to use POS relevance weighting matrices to model each word-context pair. As a result, the context word vector for predicting the central word $w_{t}$ is calculated as a weighted sum rather than a simple average operation:
\begin{equation}\label{eq:bow-vec}
  \mathbf{v}_{t-c}^{t+c} = \sum_{-c \leq i \leq c, i \ne 0} \Phi_{i}(z_{t+i}, z_{i}) \mathbf{w}_{t+i}^{(1)}
\end{equation}
where $\Phi_{i}(z_{t+i}, z_{i})$ is a core weighting factor that represents the relevance weight from POS tag $z_{t+i}$ to $z_{t}$.
The subscript $i$ indicates the position distance between word $w_{t+i}$ and $w_{t}$ within the specific training context window.
Therefore, the POS based weighting factors are \textbf{position-dependent}.

\begin{figure}[htb]
\begin{minipage}[b]{1.0\linewidth}
  \centering
  \includegraphics[width=7.5cm]{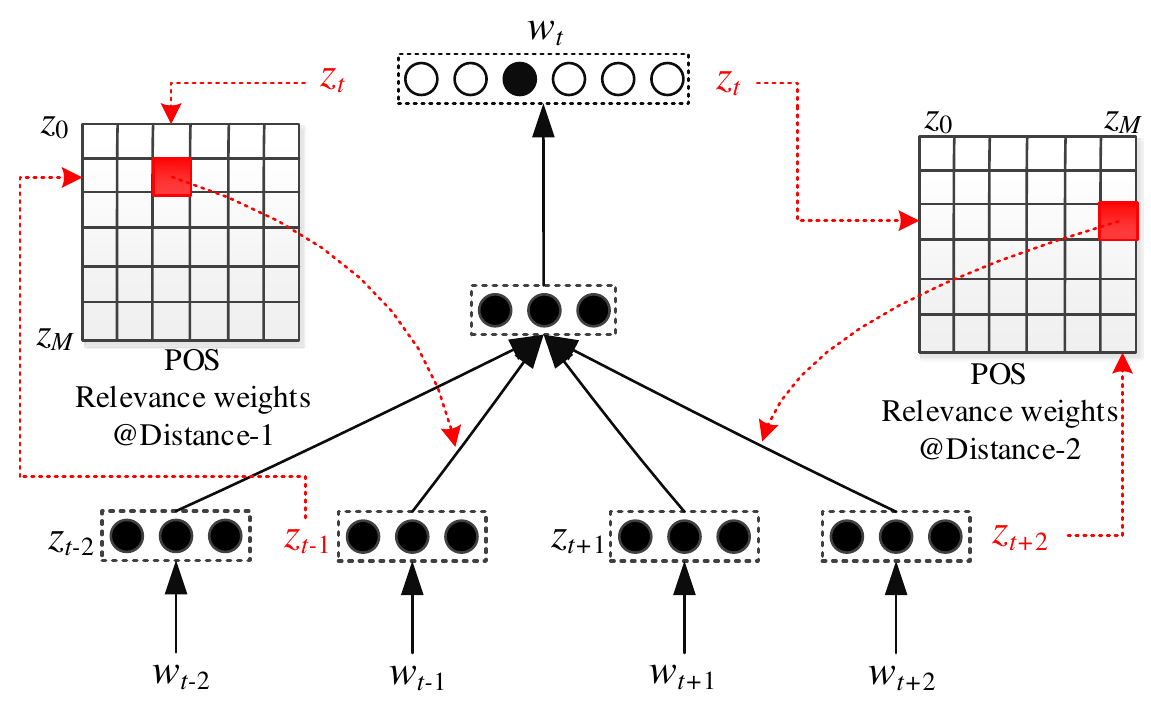}\\
\end{minipage}
\caption{CBOW based PWE framework.}
\label{fig:cbow-pwe}
\end{figure}

The main framework of the proposed POS based word embedding model is depicted in Figure \ref{fig:cbow-pwe}.
In this framework, we propose to model the word-context pair based on the relevance weights between the corresponding two POS tags.

%%%%%%%%%%%%%%%%%%%%%%%%%

For example, if we consider the central word $w_{t}$ and its context word $w_{t-1}$,
we define a POS relevance weighting matrix $\Phi_{-1}$ from which $\Phi_{-1}({z_{t-1}, z_{t}})$ is used as a weighting factor based on the corresponding POS tags $z_{t-1}$ and $z_{t}$.
Meanwhile, considering word $w_{t}$ and its context word $w_{t+2}$, since the distance between them is 2, we define another POS relevance weighting matrix $\Phi_{2}$ where again we could extract a factor $\Phi_{2}({z_{t+2}, z_{t}})$ for context weighting.

\subsection{Optimization algorithm}
\label{ssec:train}
In the PWE framework, all the POS relevance weights are updated jointly with the word representations during training.
This work proposes to apply the stochastic gradient descend (SGD) algorithm to parameter learning.
The key issue for training PWE model is to calculate the derivatives of the POS relevance weighting matrices.
The partial derivatives of the objective function with respect to word embeddings are
\begin{equation}\label{gradient-vector}
  %v_{t-c}^{t+c} = \lambda_{\mathrm{pos}} v_{\mathrm{pos}} + \sum_{\mid i \mid < C, i \ne 0}{ \lambda_{i} v_{t+i} }
  \frac {\partial {\cal Q}_{\mathrm{cbow}}} {\partial \mathbf{w}_{t+i}^{(1)}} =
  \Phi_{i}(z_{t+i}, z_{t}) \ \frac {\partial {\cal Q}_{\mathrm{cbow}}} {\partial \mathbf{v}_{t-c}^{t+c}}
\end{equation}
where $\frac {\partial {\cal Q}_{\mathrm{cbow}}} {\partial \mathbf{v}_{t-c}^{t+c}}$ could be calculated efficiently under the typical word embedding framework.
%, which is equal to $\frac {\partial {\log p(w_{t} | w_{t-c}^{t+c})}} {\partial \mathbf{v}_{t-c}^{t+c}}$.
Meanwhile, the partial derivative with respect to the POS relevance weight can be computed as
\begin{equation}\label{gradient-weight}
  %v_{t-c}^{t+c} = \lambda_{\mathrm{pos}} v_{\mathrm{pos}} + \sum_{\mid i \mid < C, i \ne 0}{ \lambda_{i} v_{t+i} }
  \frac {\partial {\cal Q}_{\mathrm{cbow}}} {\partial \Phi_{i}(z_{t+i}, z_{t})} =
  \frac {\partial {\cal Q}_{\mathrm{cbow}}} {\partial \mathbf{v}_{t-c}^{t+c}} \cdot \mathbf{w}_{t+i}^{(1)}
\end{equation}

For efficiency purpose, this paper uses the negative sampling technique during training process while the POS
tags of the negative samples is set to be the same as the tags of the training tokens.
The baseline models are trained based on \textit{dynamic context window} using the popular word2vec toolkit \cite{mikolov2013efficient}\footnote{https://code.google.com/p/word2vec/}.
%During the model training process, they using dynamic window for taking into account proximity.
In the next section, we will report our experimental results without using dynamic context window and prove the effectiveness of the proposed model.

\section{Experiments}
\label{sec:experm}
\subsection{Experimental setup}
\label{ssec:experm-setup}
%Here we conduct experiments on the popular word similarity and word analogy task.
In this section, we present the experimental setup, which including the training corpus, POS tagger and parameter settings for all experiments.

\textbf{Training Corpus}:
This paper used the April 2010 snapshot of the Wikipedia corpus \cite{shaoul2010westbury} with a total of about 2 million articles and 1 billion tokens.
The Wikipedia corpus was pre-processed
%by removing all the HTML meta-data and hyper-links and replacing the digit numbers with English words
using the perl script from the Matt Mahoney's page\footnote{http://mattmahoney.net/dc/textdata.html}.
After normalization, we constructed a vocabulary with 212,300 distinct words by discarding words that occur less than 50 times.
%Finally, the training corpus contains about 1 billion tokens.

\textbf{Part-of-speech Tagger}:
To obtain the POS tags for each training token, we used the OpenNLP toolkit\footnote{http://opennlp.apache.org/} for part-of-speech tagging.
The tag set is the Penn Treebank POS tag set \cite{marcus1993building}, which consists of 36 common POS tags and 6 symbol tags.
%The model for tagging is downloaded from the OpenNLP website, which is the standard English maximum entropy model with version 1.5.3.
%It takes about 8 hours for tagging the 1 Billion-sized corpus on a single CPU machine.

\textbf{Experiment Setting}:
%In this work, the tag set of part-of-speech refers to the Penn Treebank POS tag set \cite{marcus1993building}.
%For simplified description, those tags could be divided into 6 groups \{N, V, J, R, Other, Symbol\} as the work of \newcite{qiu2014learning}.
%``N'' includes nouns and their variations \{NN, NNS, NNP, NNPS\};
%``V'' includes verbs and their variations \{VB, VBD, VBG, VBN, VBP, VBZ\};
%``J'' includes adjectives and their variations \{JJ, JJR, JJS\};
%``R'' includes adverbs and their variations \{RB, RBR, RBS, RP\};
%``Other'' includes the rest of the POS tags \{CC, CD, DT, EX, FW, IN, LS, MD, PDT, POS, PRP, PRP\$, SYM, TO, UH, WDT, WP, WP\$, WRB\} except the ``Symbol" set, which contains symbols ``:'', ``\$'', etc.
As for word embedding training, we set the vector dimensionality to 300 and the negative sample number to 5.
The learning rate was set to 0.025 and then decreased with the training process \cite{mikolov2013efficient}.
The context window size was set to 5 for all models.
All the weights in the POS relevance weighting matrices were initialized uniformly.
For all the experimental results reported in this paper, we used just one training epoch for comparison.
Meanwhile, all the experimental results were obtained based on repeated running and averaging.

\subsection{Qualitative analysis}
\begin{figure}
  \begin{minipage}[t]{0.5\linewidth}
    \centering
    \includegraphics[width=4cm]{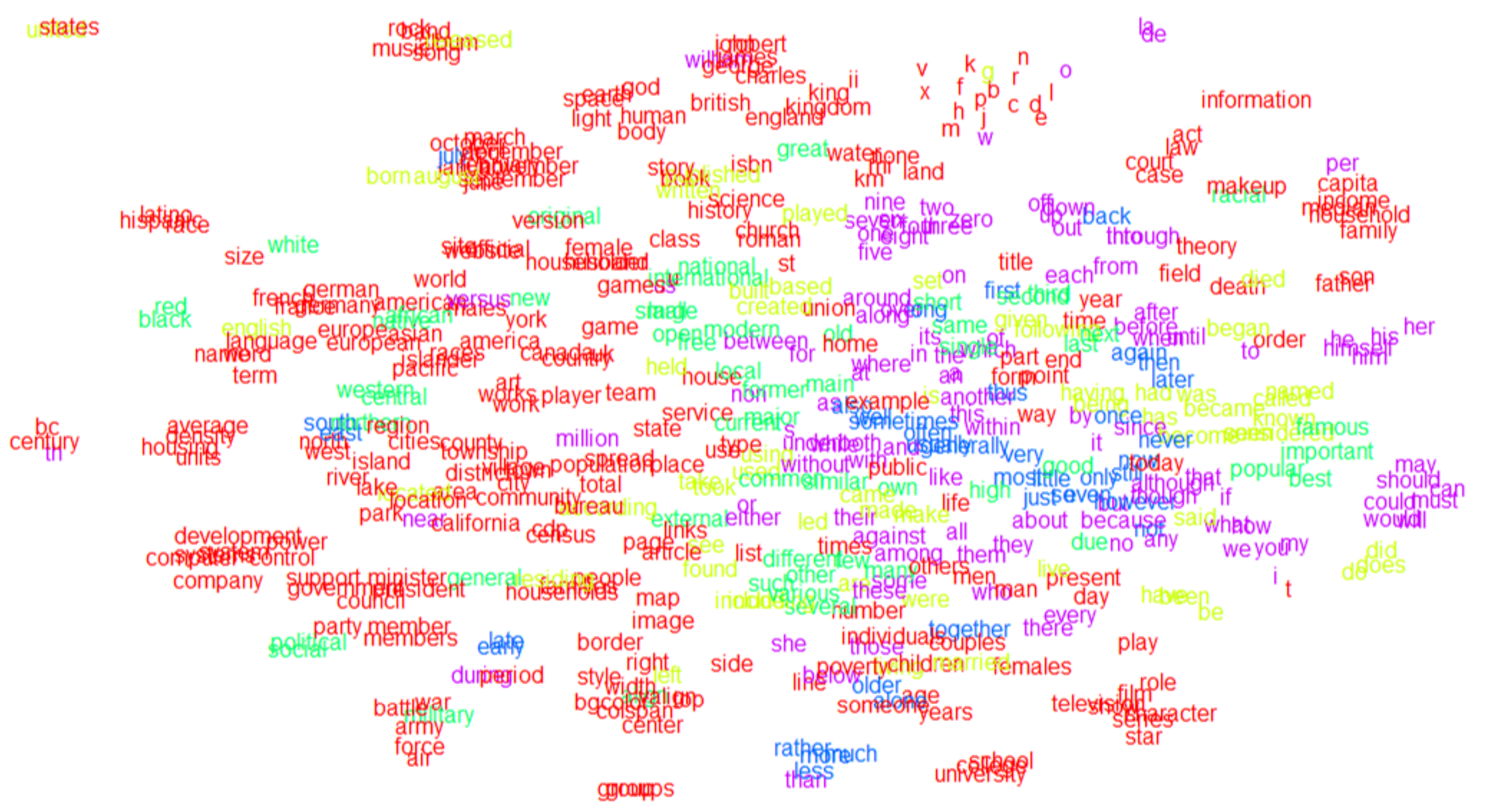}
    \caption{The CBOW model}
    \label{fig:top500-cbow}
  \end{minipage}%
  \begin{minipage}[t]{0.5\linewidth}
    \centering
    \includegraphics[width=4cm]{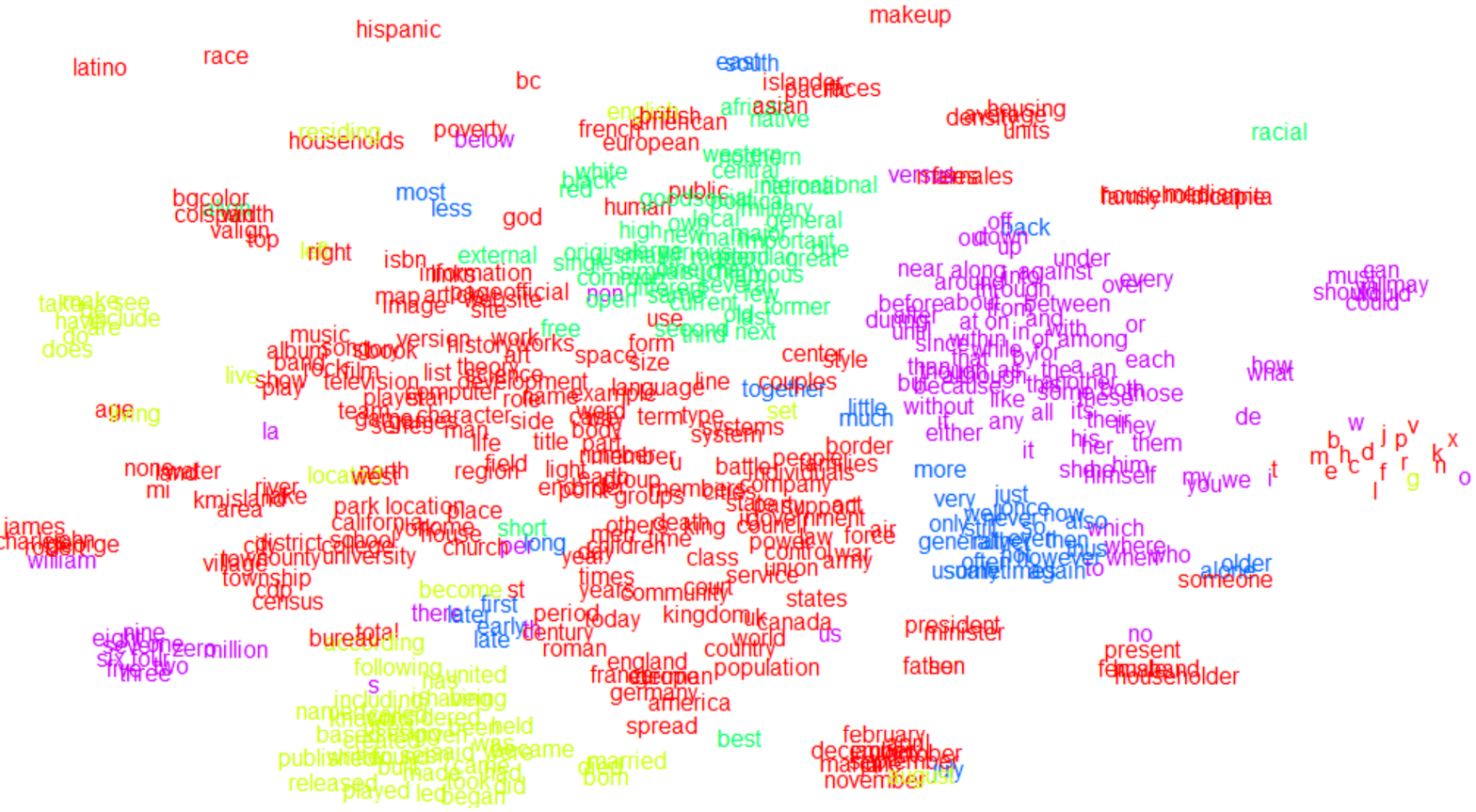}
    \caption{The PWE model}
    \label{fig:top500-pwe}
  \end{minipage}
\end{figure}

We firstly plot the top 500 words in vocabulary based on word vectors learned by the CBOW and the proposed model in Figure \ref{fig:top500-cbow} and Figure \ref{fig:top500-pwe}.
For simplified visulization, all main POS tags are divided into 5 coarse-grained groups \{N, V, J, R, Other\} as the work of \newcite{qiu2014learning}.
%``N'' includes nouns and their variations \{NN, NNS, NNP, NNPS\};
%``V'' includes verbs and their variations \{VB, VBD, VBG, VBN, VBP, VBZ\};
%``J'' includes adjectives and their variations \{JJ, JJR, JJS\};
%``R'' includes adverbs and their variations \{RB, RBR, RBS, RP\};
%``Other'' includes the rest of the POS tags \{CC, CD, DT, EX, FW, IN, LS, MD, PDT, POS, PRP, PRP\$, SYM, TO, UH, WDT, WP, WP\$, WRB\} except the ``Symbol" set, which contains symbols ``:'', ``\$'', etc.
%From the visualization results, we find the pro
The corresponding colors in the visualization figures are \{\textit{red}, \textit{canary}, \textit{green}, \textit{blue}, \textit{pink}, \textit{red}\} respectively.
We find the proposed model could group words with similar POS category.
We also test our model based on unsupervised POS induction for the top 500 words \cite{christodoulopoulos2010two,yatbaz2012learning}.
By using the k-means clustering algorithm, we map each cluster to the gold standard tag that is most common for the words in that cluster.
The \textit{cluster purity} is improved from 53.9\% to 74.3\% when comparing the baseline CBOW model and the proposed model.
The qualitative analysis indicates that the proposed model has the ability to model the syntactic relationships of word meanings.
%The results are given in Table \ref{tab:result-pos-induce}.
%%%%%
%\begin{table}[htb]\centering
%\begin{tabular}{|l|c|}
%  \hline
%  Model & \textit{cluster purity} \\\hline
%  CBOW &  53.9\\\hline
%  \bf PWE & 74.3 \\\hline
%\end{tabular}
%\caption{Results of POS induction (\%).}
%\label{tab:result-pos-induce}
%\end{table}

\subsection{Syntactic word analogy task}
Here we use two syntactic word analogical reasoning datasets for experiments.
The first one named as MSR is a dataset that contains 8000 morpho-syntactic analogy questions \cite{mikolov2013linguistic}.
%This dataset contains 8869 semantic questions (SEM) and 10675 syntactic questions (SYN).
The other one named as SYN is a dataset proposed in \newcite{mikolov2013efficient}, which contains 10675 syntactic questions.
To answer those analogy questions, we firstly remove all out-of-vocabulary words\footnote{This removed 1574 instances from the MSR dataset and 66 instances from the SYN dataset.} and then
use the typical similarity multiplication method to find the correct answers from the entire vocabulary \cite{levy2014linguistic}.
Our experimental results are shown in Table \ref{tab:result-analogy}.
% as well as the state-of-the-art performance
\begin{table}[htb]\centering
\begin{tabular}{|l|c|c|c|}
  \hline
  \multicolumn{2}{|l|}{Model}   & MSR & SYN \\\hline\hline
  %\multicolumn{2}{|l|}{Mikolov et al. (2013a)} & -- & 65.10 \\ \hline
%  \multicolumn{2}{|l|}{Levy et al. (2014)} & 53.98 & -- \\ \hline
  %\multirow{3}[3]*{Our Run} & CBOW & 52.16 & 63.24 \\ \cline{2-4}
%  & Skip-gram & 49.55 & 60.45 \\ \cline{2-4}
%  & \textbf{PWE} & \textbf{59.59} & \textbf{65.90} \\ \hline
  \multicolumn{2}{|l|}{CBOW} & 52.16 & 63.24 \\ \cline{2-4}
  \multicolumn{2}{|l|}{Skip-gram} & 49.55 & 60.45 \\ \cline{2-4}
  \multicolumn{2}{|l|}{\textbf{PWE}} & \textbf{59.59} & \textbf{65.90} \\ \hline
\end{tabular}
\caption{Results on syntactic analogy tasks (\%).}
\label{tab:result-analogy}
\end{table}

The experimental results on the two syntactic word analogy tasks have shown that, the proposed PWE model achieves significant improvements on the syntactic word analogy tasks. Using POS relevance weights for context weighting improves the representation of syntactic relationships in the learned word vectors.

\subsection{Word similarity tasks}
We test our models on WordSim-353 \cite{finkelstein2001placing}, MEN \cite{bruni2012distri}, and MC \cite{miller1991contextual} tasks of word similarity.
All the experimental results are given in Table \ref{tab:result-wordsim}.
%with state-of-the-art performances \cite{levy2014neural,pennington2014glove}
%The experimental results are shown in Table \ref{tab:result}.
\begin{table}[htb]\centering
\begin{tabular}{|l|c|c|c|c|}
  \hline
  % after \\: \hline or \cline{col1-col2} \cline{col3-col4} ...
  %\multirow{2}[2]*{\textit{Model}} & \multicolumn{3}{|c|}{Word similarity} & \multicolumn{2}{|c|}{Word analogy} \\
  %\cline{2-6}
  \multicolumn{2}{|l|}{Model}  & WS353 & MEN & MC \\\hline\hline
  %\multicolumn{2}{|c|}{State-of-the-art} & 68.70 & 72.10 & 72.70 \\ \hline
  %\multicolumn{2}{|l|}{SPPMI} & 68.70 & 72.10 & -- \\ \hline
%  \multicolumn{2}{|l|}{GloVe} & -- & -- & 72.70 \\ \hline
 % \multirow{3}[3]*{Our Run} & CBOW & 67.07 & 74.10 & 73.75 \\ \cline{2-5}
%  & Skip-gram & 70.29 & 73.93 & 78.51 \\ \cline{2-5}
%  & \textbf{PWE} & \textbf{71.18} & \textbf{75.43} & \textbf{81.87} \\ \hline
  \multicolumn{2}{|l|}{CBOW} & 67.07 & 74.10 & 73.75 \\ \cline{2-5}
  \multicolumn{2}{|l|}{Skip-gram} & 70.29 & 73.93 & 78.51 \\ \cline{2-5}
  \multicolumn{2}{|l|}{\textbf{PWE}} & \textbf{71.18} & \textbf{75.43} & \textbf{81.87} \\ \hline
\end{tabular}
\caption{Results on word similarity tasks (\%).}
\label{tab:result-wordsim}
\end{table}

Experimental results shown in Table 1 indicate that the proposed PWE model achieves consistent
improvements in three word similarity tasks.
This proves that the incorporation of POS relevance weights for discriminatively context weighting is useful for better modeling the word-context patterns, so as to word embeddings.

\section{Conclusion}

This paper has proposed a model for learning distributed word representations with context weighting based on POS relevance weights.
This paper designs position-dependent POS relevance weighting matrices for weighting word-context pairs during the training process of word embeddings.
In the proposed model, word embeddings and the POS relevance weighting matrices are jointly learned using the stochastic gradient descend algorithm.
Experiments conducted on the popular word analogy and word similarity tasks have all demonstrated the effectiveness of the proposed method.

%\section*{Acknowledgments}
%Do not number the acknowledgment section.

\bibliography{naaclhlt2016}
\bibliographystyle{naaclhlt2016}

\end{document}